\documentclass[a4paper,conference]{IEEEtran}
\usepackage{graphicx}
\usepackage{subfigure}
\usepackage{amssymb}
\usepackage{color}
\usepackage{latexsym}
\usepackage{float}
\usepackage{cite}
\usepackage{booktabs}
\usepackage{multirow}
\usepackage{color}
\usepackage{amsmath}
\usepackage{etoolbox}
\usepackage{hyperref}
\makeatletter
\patchcmd{\@makecaption}
  {\scshape}
  {}
  {}
  {}
\makeatother

\newcommand{\minisection}[1]{\vspace{0.04in} \noindent {\bf #1}\ \ }

\begin{document}
%

\title{Weakly Supervised Domain-Specific \\Color Naming Based on Attention}

\author{\IEEEauthorblockN{Lu Yu $^{1,2}$, Yongmei Cheng $^{1}$, Joost van de Weijer $^{2}$}
\IEEEauthorblockA{1.Key Laboratory of Information Fusion Technology, School of Automation, Northwestern Polytechnical University, Xi'an, China\\ 2. Computer Vision Center, Universitat Autonoma de Barcelona, Barcelona, Spain\\ Email: luyu@cvc.uab.es, chengym@nwpu.edu.cn, joost@cvc.uab.es}}

\maketitle
\begin{abstract}
The majority of existing color naming methods focuses on the eleven basic color terms of the English language. However, in many applications, different sets of color names are used for the accurate description of objects. Labeling data to learn these domain-specific color names is an expensive and laborious task. Therefore, in this article we aim to learn color names from weakly labeled data. For this purpose, we add an attention branch to the color naming network. The attention branch is used to modulate the pixel-wise color naming predictions of the network. In experiments, we illustrate that the attention branch correctly identifies the relevant regions. Furthermore, we show that our method obtains state-of-the-art results for pixel-wise and image-wise classification on the EBAY dataset and is able to learn color names for various domains.
\end{abstract}


\section{Introduction}
\label{sec:intro}

Color is a basic characteristic of visual objects in the world. As one of the important features of visual data, colors are crucial in understanding the world, and they can be used to distinguish one object from another in our daily life. Humans use color names to refer to a specific color and to communicate color information with other humans. Examples of color names are 'blue', 'crimson' and 'amber'.  Computational color naming aims to identify color names in images; this is usually done by learning a mapping between color values and color names. Computational color naming is important for applications in human computer interaction, including online shopping, fashion analysis, image retrieval and person re-identification \cite{liu2014fashion,cheng2016pedestrian,yu2018beyond}.

For the purpose of this article, we divide computational color naming models in methods which are trained in a supervised or semi-supervised manner. Supervised methods are based either on labeled color patches \cite{Benavente2012,mojsilovic2005computational} or on pixel ground-truth masks, providing the color names for all the relevant items in the image \cite{liu2014fashion,cheng2016pedestrian}. The work of Van de Weijer et al. \cite{van2009learning} proposed a method to learn color names from images retrieved from Google in a semi-supervised manner.  We refer with {\it semi-supervised} to the fact that the provided label describes the color of the principal object in the image, but no information on the exact pixels which are described by the label is given. An advantage of semi-supervised methods is that they reduce the label effort significantly. However, the existing unsupervised methods \cite{liu2014fashion,cheng2016pedestrian,van2009learning,wang2015color,yuan2015illumination} still require pixel masks at the testing phase. The methods are therefore semi-supervised at training, but supervised at test time.

\begin{figure}[tb]
\centering
\subfigure[]{\includegraphics[width=33mm]{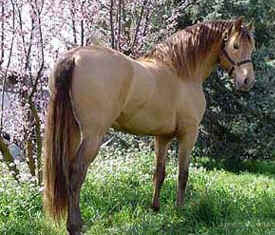}}
\subfigure[]{\includegraphics[width=18.5mm]{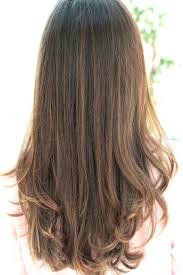}}
\subfigure[]{\includegraphics[width=34.5mm]{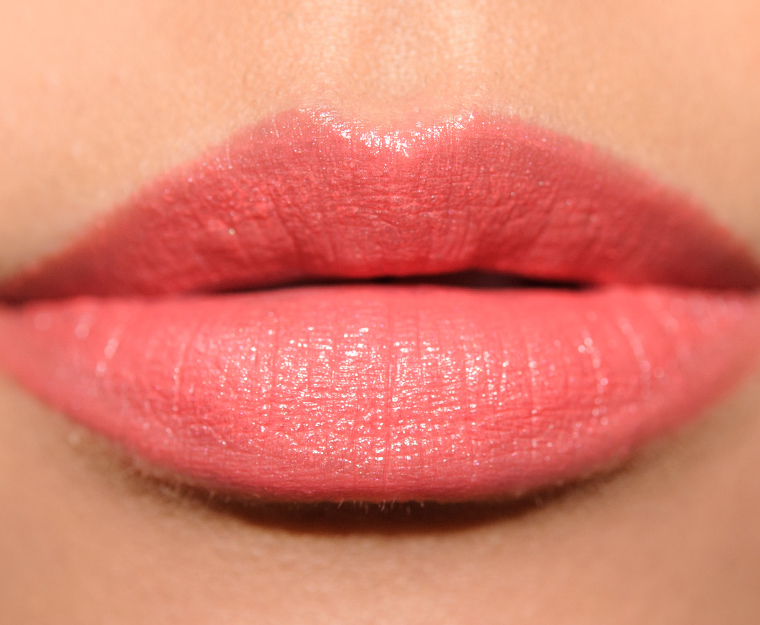}}
\caption{ Example images of domain-specific color names: (a) 'champagne' colored horse,  (b) 'almond' colored hair and (c) 'coral red' lips.} \label{fig:examples}
\end{figure}

The vast majority of the existing color naming approaches use the eleven basic color terms~\cite{mojsilovic2005computational,van2009learning,yuan2015illumination} which were defined in the seminal study of Berlin and Kay~\cite{hardin1997color}. Even though these color names are widely used, many applications apply different domain-specific color names. In Fig.~\ref{fig:examples} several examples of color names within various applications are provided: a 'champagne' colored horse, 'almond' colored hair and 'coral red' lips. Because different application domains use different sets of color names, the laborious labeling process would need to be performed repeatedly. Therefore, in this paper we aim for a method which can learn from weakly labeled data, and which does not require any supervision at testing time. Learning from weakly labeled data has been studied before for image classification \cite{mnih2012learning,xiao2015learning}, image segmentation \cite{pathak2015constrained,joon2017exploiting}, saliency detection \cite{wang2017learning}, object detection \cite{chen2015webly,bilen2015weakly}, and object recognition \cite{fergus2010learning,krause2016unreasonable,niu2015visual,wang2017multi}. 

\begin{figure*}[t]
\begin{center}
  \includegraphics[width=0.85\textwidth]{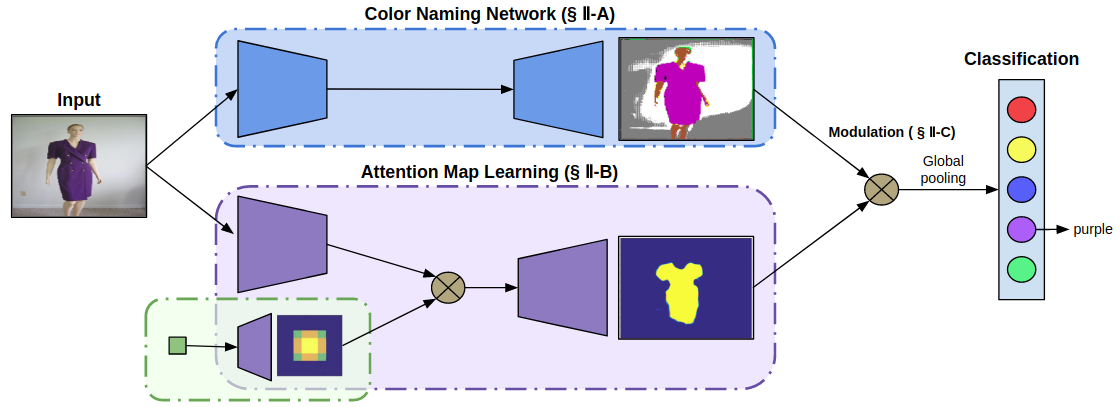}
  \caption{Overview of our proposed framework for weakly supervised color name prediction. Our model is capable of automatically discovering correct regions of interest for image-wise color label predicting and simultaneously providing an end-to-end mapping between color values and color names.
} \label{fig:framework}
\end{center}
\end{figure*}

To address the drawback of the deep learning approaches for color naming, we propose a weakly-supervised deep learning framework based on attention. The main contribution of our paper is a new two-branch network design for color naming based on attention, which is capable of automatically discovering relevant regions related to weak image labels, and simultaneously learn a mapping between color values and color names. In addition, we collect a large-scale dataset using a Web image search engine, which contains 11 basic color naming images for 4 categories, and a dataset for domain-specific color naming which includes color names for horses, eyes colors, lips colors, and the tomato growing stages. Experiments show that our attention network correctly identifies the relevant image regions, and at the same time learns a mapping from image values to color names.

\section{Attention Modulation for Color Naming}
\label{sec:CNN}

The aim of this article is to predict the color name that best describes the principal object in the image. The method is to be trained from weakly-labeled data, which means  a color name label is provided for the image, but that no segmentation mask or bounding box is provided to identify the principal object. We assume that images contain a single principal object which can be described by a single color name. 

To train from the weakly-supervised data the method has to perform two tasks: identify the principal object in the scene and predict the color name which best describes its color. In the design of the network, which is provided in Fig.\ref{fig:framework}, we have two parallel branches, one for each task. The first branch is a shallow convolutional neural network which aims to predict a pixelwise color name map. The second branch computes an attention map which identifies the regions which contain the relevant color name information. The two branches are combined with a modulation part which combines the automatically learned attention map with each channel of the predicted color naming map.

\subsection{Color Naming Network (CN-CNN)}
\label{sec:CN-CNN}
\begin{figure}[tb]
\begin{center}
  \includegraphics[width=0.45\textwidth]{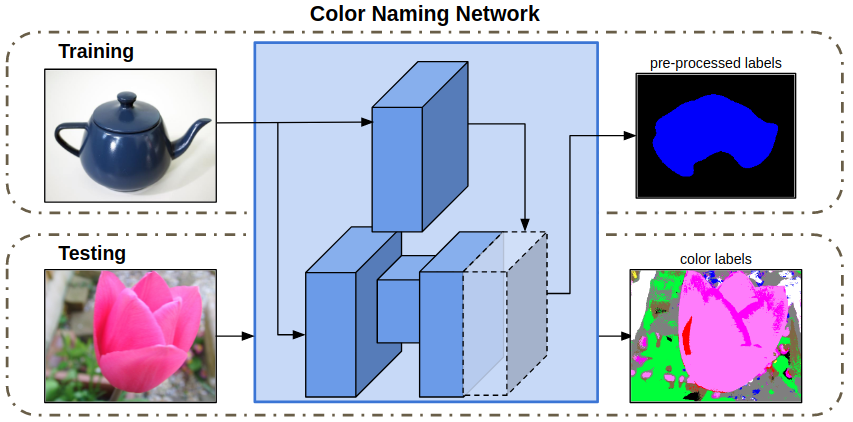}
  \caption{The structure of the color naming network (CN-CNN)} \label{fig:step1}
\end{center}
\end{figure}

The color naming network takes a color image $I\in \mathbb{R}^{H\times W\times 3}$ as an input and produces an estimate of the color name distribution $Y\in \mathbb{R}^{H\times W\times C}$ where $C$ is the number of color names. The structure of the CN-CNN is illustrated in Fig.~\ref{fig:step1} (see also top row of Table~\ref{table:details}). Specifically, first it passes through several convolutional and pooling layers after which we apply deconvolution to arrive back at the original image size. Then the features after one convolutional layer from the original input are concatenated to the part after the deconvolution layer with a skip layer \cite{shelhamer2017fully}. One soft-max layer is then added to normalize the distribution of all dimensions.

\begin{table*}[tb]
\centering
\caption{Details of our network}
\label{table:details}
\tabcolsep=0.03cm
\begin{tabular}{|c|c|c|c|c|c|c|c|c|c|c|c|}
\hline
\multirow{4}{*}{CN-CNN} & Type               & Conv+BN+Relu   & Maxpool & Deconv+BN+Relu & Conv+BN+Relu & Concat     & Conv           & Softmax   &            &         &         \\ \cline{2-12} 
                        & Filters            & 72             &         & 72             & 72           &            & 11             &           &            &         &         \\
                        & Stride or Upsample & 1*1            & 3*3 / 2 & 3*3 / 2        & 1*1          &            & 1*1            &           &            &         &         \\
                        & Output             & 227*227        & 113*113 & 227*227        & 227*227      &            & 227*227        &           &            &         &         \\ \hline
\multirow{4}{*}{VA-CNN} & Type               & FCN-8S  (1-31) & FC+Relu & FC+Relu        & Deconv       & Modulation & FCN-8S (36-43) & Conv+Relu & Modulation & Avepool & Softmax \\ \cline{2-12} 
                        & Filters            &                & 512     & 512            & 1            &            &                & 1         &            &         &         \\
                        & Stride or Upsample &                & 7*7     & 1*1            & 8*8 / 4      &            &                & 3*3       &            & Global  &         \\
                        & Output             &                & 8*8     & 8*8            & 8*8          &            &                & 227*227   &            & 1*1     &         \\ \hline
\end{tabular}
\end{table*}

Before training the CN-CNN in an end-to-end fashion, we initialize the network by using the weak-labels of the images. We train the CN-CNN by minimizing a weighted cross entropy loss: 

\begin{equation}
L = \sum\limits_i {\sum\limits_y {\sum\limits_x {m_i \left( {x,y} \right)\log Y_i \left( {x,y,l_i } \right)} } } 
\end{equation}
where the summations are over the spatial coordinates $x$ and $y$ and image indexes $i$, and $l_i$ is the ground truth label of image $i$ and $m_i$ is a mask. Since not all the pixels in the image are correctly described by the weak-label of the image, we use a mask which is computed with a standard saliency algorithm~\cite{harel2007graph} that has a value of one for the salient part of the image and zero otherwise. This mask provides a very rough estimate of the important parts of the image, but we found this to be sufficient to provide an initialization of the network. Note that this loss is not used when training end-to-end with the whole two-branch network.

\subsection{Visual Attention Network (VA-CNN)}

Direct training on images with only the weak-labels is expected to lead to unsatisfactory performance. To further improve the visual attention network (VA-CNN),  which should identify the relevant parts of the principal object in the image. To obtain this, we propose to use an attention network branch (in purple in Fig.~\ref{fig:framework}). This branch has a color image as input and aims to compute an attention map $A\in \mathbb{R}^{H\times W\times 1}$ as output. The architecture of the attention network is based on the popular fully convolutional semantic segmentation network FCN-8s\cite{shelhamer2017fully} followed by one ReLu layer (see for details Table \ref{table:details}). The final output of the network provides the importance of each pixel for the task of color naming.

One drawback of FCN is that it cannot learn the spatial prior. However, the principal salient object in the image is most likely in the center of the image~\cite{judd2009learning}. We therefore add a spatial prior layer into the visual attention network. This layer exists of a single pixel input with value  equal to one, followed by a deconvolution layer which outputs the spatial prior. This spatial prior is then used to modulate the downsampled features of the FCN network, in the same way as the modulation layer which is explained in the following section. By backpropagating, the weights of this deconvolutional layer learns the spatial prior of the dataset.

Attention model have been applied  in various network architectures. They are used
to attend to the relevant region in the image related to text in captioning~\cite{xu2015show} or visual question answer networks~\cite{xu2016ask}. Also they have been studied in 
various computer vision tasks, including image recognition \cite{wang2017multi},  and saliency detection \cite{kuen2016recurrent}.

\subsection{Modulation Layer}

In this weakly supervised learning system, neither the ground-truth color names of each pixel nor the ground-truth of the confidence map is provided for directly training the CN-CNN or VA-CNN. We therefore propose an indirect method to jointly train both branches with only weak-labels. For this purpose  the pixel wise color name predictions $Y$ and the visual attention map $A$ with a modulation layer to output the final color name prediction for the image $Z\in \mathbb{R}^{C}$. The modulation layer does a channel wise multiplication of the feature maps of $Y$  with the attention map $A$ according to

\begin{equation}
\hat Y_k \left( {x,y,l_i } \right) = A\left( {x,y} \right)Y_k \left( {x,y,l_i } \right)
\end{equation}
where $Y_{k}$ denotes the $k$-th channel with $k= \left \{ 1, \cdot \cdot \cdot , C \right \}$; $A$ is the attention map; Score aggregation is then performed on $\hat{Y}$ using  average pooling to predict image-level score $\hat{y}$ for the $k$-th category. 

The back propagation for the modulation layer is as follows:
\begin{equation}
\frac{ \partial ( \hat{Y})}{ \partial (Y_{k})} = A  
\end{equation}

\begin{equation}
\frac{  \partial ( \hat{Y}) }{\partial (A) }= \sum_{k=1}^{C}Y_{k}\end{equation}

\subsection{Network Training}
Both CN-CNN and VA-CNN can be trained by minimizing the cross entropy loss $L\left ( l,\hat{y} \right )$, where $l$ is the ground truth label. We found that it was difficult to train the network jointly, and therefore propose an alternating training scheme. Specifically, after the CN-CNN is trained, we fix this part and fine-tuning the VA-CNN to learn attention map. After several epochs we stop training the VA-CNN branch and freeze it, and change to train the CN-CNN part again, and we repeat this process till the loss converges.



\section{Color Naming Data Collection}
\label{Sec:dataset}

\begin{figure}[tb]
\begin{center}
  \includegraphics[width=0.5\textwidth]{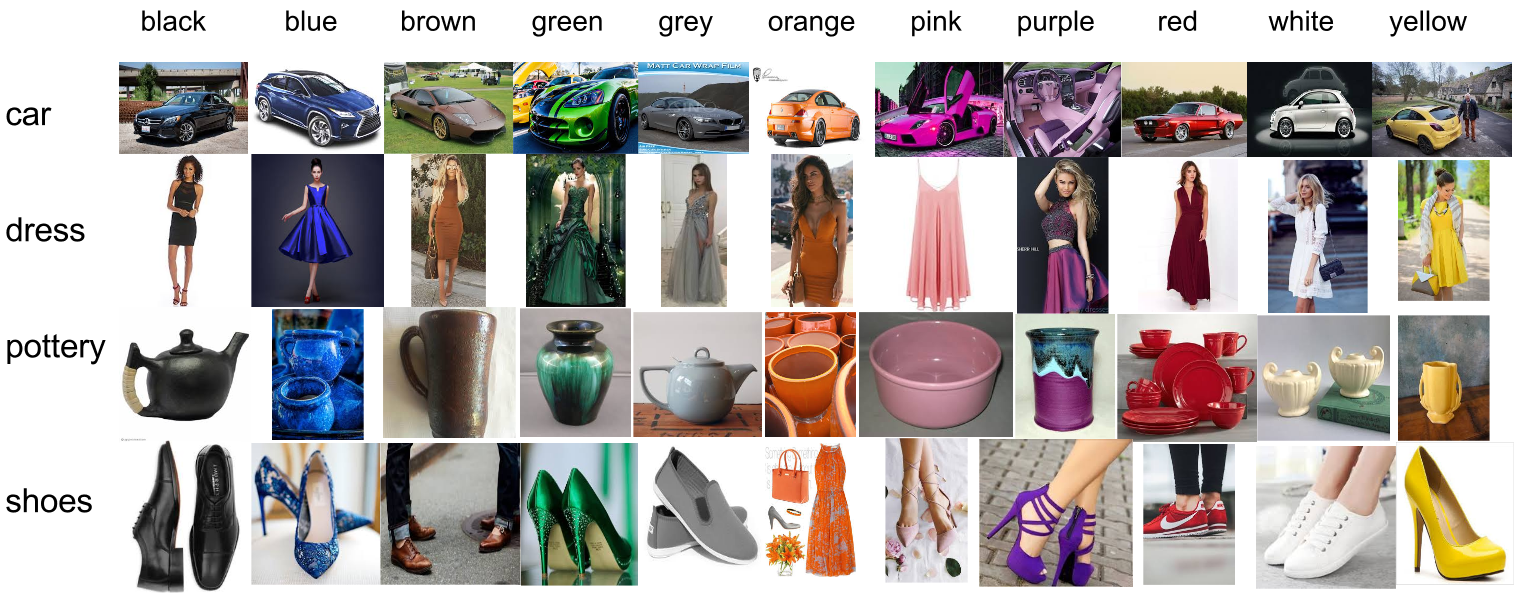}
  \caption{Examples of four categories ('car','dress','pottery' and 'shoes') in eleven basic color are shown.} \label{fig:dataset}
\end{center}
\end{figure}

\begin{figure*}[tb]
\begin{center}
  \includegraphics[width=0.85\textwidth]{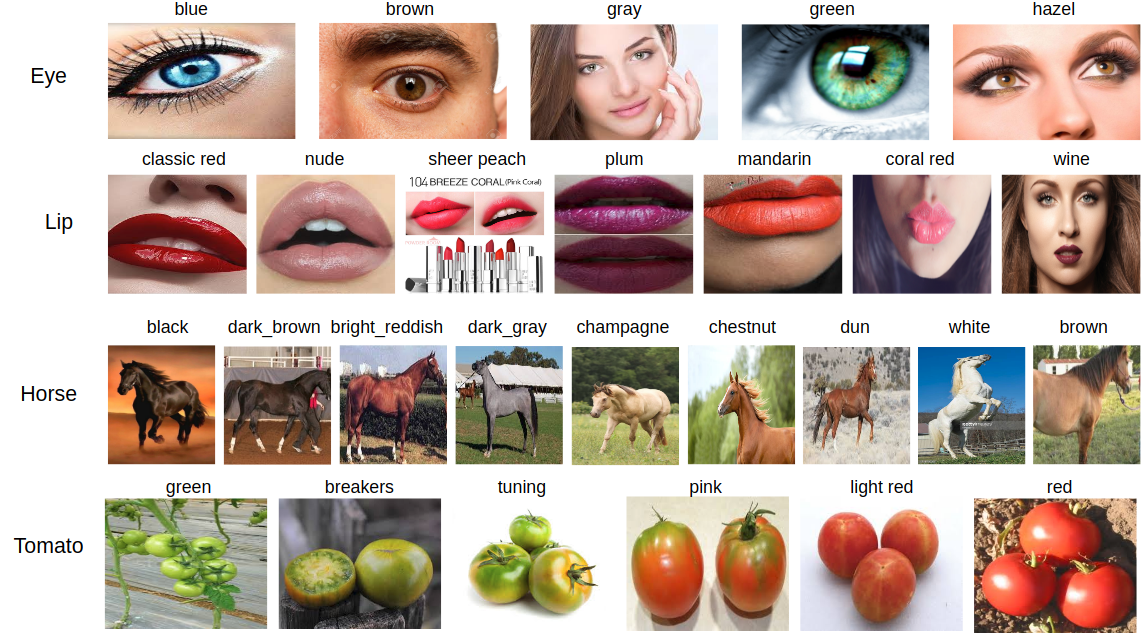}
  \caption{Examples from domain-specific datasets. One example for each domain-specific color name is shown.} \label{fig:domain}
\end{center}
\end{figure*}

For the purpose of this paper we collect two datasets: one of domain-specific color names, and one class-specific basic color term dataset. Both datasets are weakly-labeled. 

\minisection{Domain-specific color name dataset:} we collect several domain-specific datasets from Google search engine by using the query of 'color name + object': 5 colors of eyes, 7 colors of lips, 9 colors of horses and tomatoes in 6 growing stages. Then, we manually removed the noisy images. Each class has 40 images for training, 10 for validation and 20 for testing. In total, 50 images for each class of each group for domain-specific color naming learning. The dataset is available at \url{https://github.com/yulu0724/AttentionColorName}. Examples are shown in Fig.~\ref{fig:domain}. 

\minisection{Class-specific basic color term dataset:}
Since existing methods report on the eleven based color terms we also collect a class-specific dataset for these color names. We collected 2200 images from Google Image on-line by using the query of 'color name'+'object'. We choose the 11 basic color names as the indicated in \cite{berlin1991basic}, the difference is that four specific categories 'car','dress','pottery' and 'shoes' are selected as our 'object' class (the same categories as the EBAY color name dataset in Section~\ref{sec:ebay}) to decrease the probability of false positives, and adapt to our method. Hence for red, the query is 'red+car', 'red+dress', 'red+pottery' and 'red+shoes'. Then, we manually removed the noisy images. We retrieve 50 images for each color and object, so 200 images in total for each color name.
Four special categories examples for the 11 color names are given in Fig.\ref{fig:dataset}.


\section{Experiments}
\label{sec:Exp}

\subsection{Implementation Details}
We implemented our method with Matconvnet framework. 
The CN-CNN part is first pre-trained using the saliency method \cite{harel2007graph} to get a rough mask of the principal object as explained in Section \ref{sec:CN-CNN}. Next we perform alternating training of the two branches using the weakly labeled data. The newly added layers in our network are initialized with Xavier method. All the training images are resized to $227\times 227$ in our experiments for both of the CN-CNN and VA-CNN. Both of the models are optimized using Stochastic Gradient Descent (SGD) method with a batch size of 32 and 6 respectively, and a momentum of 0.9. The learning rate is set to 0.01 initially and divided by 10 after 20 epochs.


\begin{table*}[tb]
\begin{center}

\caption{Comparison of state-of-the-art methods, testing on the EBAY dataset, training with class-agnostic dataset and new class-specific dataset. We indicate with test type which methods are supervised (S) or unsupervised (U).}
\label{table: all}
\begin{tabular}{c|c|c|ccccc|ccccc}
\hline
\multirow{2}{*}{dataset}                       &                     &                   & \multicolumn{5}{c|}{pixel\_wise}                   & \multicolumn{5}{c}{image\_wise}                   \\ \cline{2-13} 
                                & Method              & test type & car   & dress & shoes & pottery & \textbf{overall} & car   & dress & shoes & pottery & \textbf{overall} \\ \hline
\multirow{4}{*}{class-agnostic dataset} & PLSA                & S              & 56.00 & 80.00 & 77.00 & 70.00   & \textbf{70.60}   & 74.00 & 85.00 & 94.00 & 82.00   & \textbf{83.40}   \\
                                & SS\_net             & S              & -     & -     & -     & -       & \textbf{74.00}   & 73.18 & 91.82 & 91.18 & 83.36   & \textbf{84.89}   \\
                                & Ours     & S             & 51.38 & 80.27 &  77.64 & 71.03   & \textbf{71.83}   & 71.32 & 86.36 & 88.18 & 80.91   & \textbf{81.82}   \\
                                & Ours  & U             & -     & -     & -     & -       & \textbf{-}       & 63.64 & 79.07 & 81.82 & 74.55   & \textbf{74.77}   \\ \hline
\multirow{4}{*}{class-specific dataset} & PLSA                & S              & 54.52 & 82.75 & 75.37 & 71.98   & \textbf{71.15}   & 69.09 & 93.64 & 89.09 & 87.27   & \textbf{84.77}   \\
                                & Classification      & U             & -     & -     & -     & -       & \textbf{-}       & 66.36 & 78.18 & 70.91 & 72.73   & \textbf{72.05}   \\
                                & Ours     & S             & 57.88 & 85.35 & 78.32 & 75.54   & \underline{\textbf{74.27}}   & 73.64 & 94.55 & 94.55 & 86.36   & \underline{\textbf{87.27}}   \\
                                & Ours & U             & -     & -     & -     & -       & \textbf{-}       & 72.72 & 94.54 & 84.55 & 87.27   & \underline{\textbf{86.59}}   \\ \hline
                                & Human               & -                 & -     & -     & -     & -       & \textbf{-}       & -     & -     & -     & -       & \textbf{88.98}   \\ \hline
\end{tabular}
\end{center}
\end{table*}

\begin{figure}[tb]
\centering
\subfigure[]{\includegraphics[width=40mm]{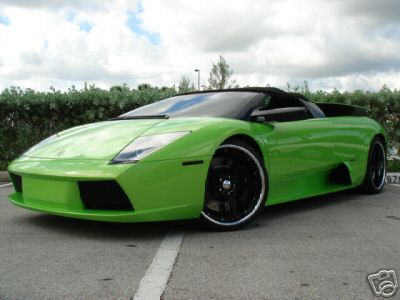}}
\subfigure[]{\includegraphics[width=40mm]{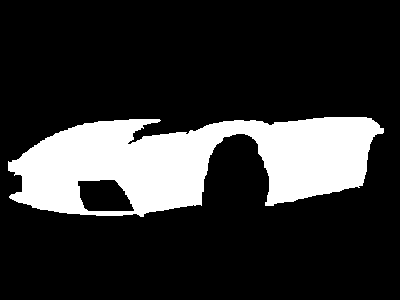}}
\caption{ (a) Example image from EBAY labeled with the color name 'green' and (b) the ground truth mask of the image identifying the pixels which are related with the color name. } \label{fig:car}
\end{figure}

\subsection{Color Naming from Weakly Labeled Data}
\label{sec:ebay}
Most existing methods on color naming are trained with the eleven basic color terms. We start with an ablation study to evaluate our  method, and next compare it to other methods. 

We compare results on the EBAY dataset which contains a total of 440 images, consisting of ten images for the eleven color names for four different categories (cars, shoes, dresses, and pottery). All images come with a mask image which identifies the pixels which belong to the named object. This Evaluation is only performed for the pixels in the mask, see example in Fig.~\ref{fig:car}.

\minisection{Ablation Study:} We perform an ablative study to analyze the contribution of the critical components of our proposed model. The results are on our class-specific dataset and summarized in Table~\ref{table:ablation}. They show a drop of about 2\% without applying alternating learning, 2.5\% drop without further adding centric information, and a significant drop when removing  all of these, which demonstrates the relevance of the components we propose.

\begin{table}[tb]
\centering
\caption{Comparison of our model learned using different components on the EBAY dataset. We abbreviate attention, centric information and alternating learning as AM, C, AL.}
\label{table:ablation}
\begin{tabular}{c|c}
\hline
                                                                 & Accuracy              \\ \hline
Ours                                                             &   55.45                   \\
Ours+AM                                                   &   84.09                   \\
Ours+AM+C                                           &   84.77                   \\
Ours+AM+C+AL                      & 86.59                 \\ \hline
\end{tabular}
\end{table}

\begin{table*}[tb]
\centering
\caption {Color naming results on Eye, Lip, Horses and Tomato dataset respectively comparing to using classification network (pre-trained AlexNet)}
\label{table:domain1}
\begin{tabular}{c|cccccccccc|c}
\hline
Dataset  & \multicolumn{10}{c|}{Ours}                                                                                                   & Classification   \\ \hline
Eye  & blue         & brown        & gray            & green      & hazel     &          &       &       &       & \textbf{overall} & \textbf{overall} \\
Accuracy & 65.00        & 85.00        & 65.00           & 70.00      & 10.00     &          &       &       &       & \textbf{59.00}   & \textbf{49.00}   \\ \hline
Lip      & classic\_red & sheer\_peach & coral\_red      & mandarin   & nude      & plum     & wine  &       &       & \textbf{overall} & \textbf{overall} \\
Accuracy & 65.00        & 40.00        & 55.00           & 70.00      & 60.00     & 35.00    & 65.00 &       &       & \textbf{55.72}   & \textbf{45.00}   \\ \hline
Horse    & black        & dark\_brown  & bright\_reddish & dark\_gray & champagne & chestnut & dun   & white & brown & \textbf{overall} & \textbf{overall} \\
Accuracy & 80.00        & 45.00        & 15.00           & 85.00      & 80.00     & 70.00    & 30.00 & 90.00 & 45.00 & \textbf{60.00}   & \textbf{58.89}   \\ \hline
Tomato   & green        & breakers     & tuning          & pink       & light red & red      &       &       &       & \textbf{overall} & \textbf{overall} \\
Accuracy & 55.00        & 25.00        & 60.00           & 35.00      & 65.00     & 80.00    &       &       &       & \textbf{53.33}   & \textbf{50.83}   \\ \hline
\end{tabular}
\end{table*}

\begin{figure}[tb]
\begin{center}
  \includegraphics[width=0.48\textwidth]{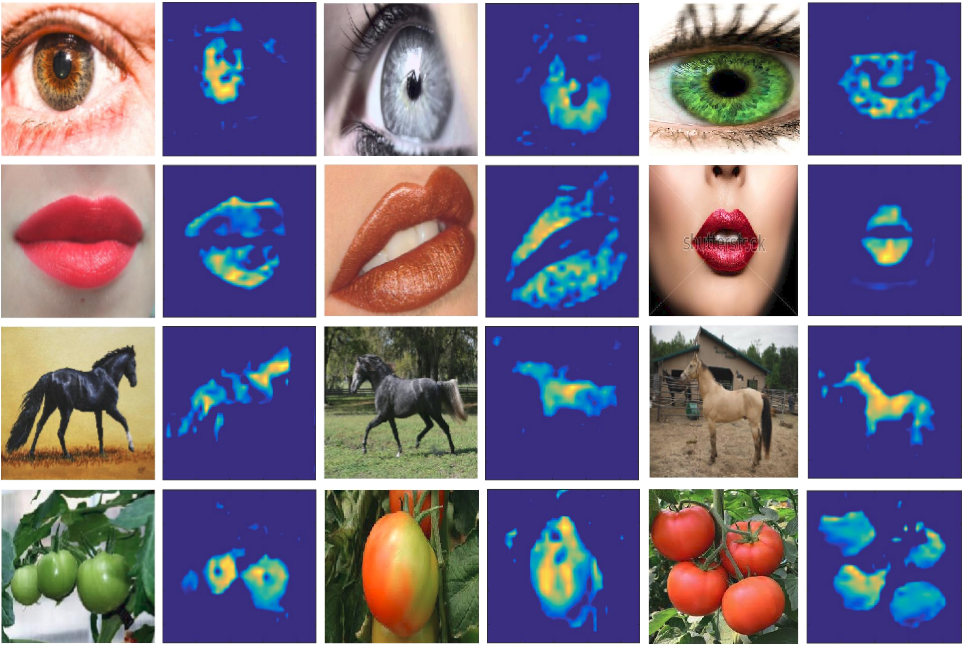}
  \caption{Examples of Attention map from Eye, Lip, Horse and Tomato datasets.} \label{fig:domain_map}
\end{center}
\end{figure}

\minisection{Comparison with the State-of-the-art:} In Table~\ref{table: all} we compare our results testing on the EBAY dataset with previous related work: PLSA~\cite{van2009learning}, SS\_{net}~\cite{wang2015color} with different training data. All methods train from weakly labeled data, however it is important to stress that only our method can be applied unsupervised at test time, while the other methods need a mask of the object (this is indicated with U and S in Table~\ref{table: all}). We provide results for pixel-wise accuracy which is defined as the percentage of correctly classified pixels, and image-wise accuracy which is defined as the percentage of images which is correctly labeled. For the pixel-wise accuracy we only use the CN-CNN network.

When comparing the methods based on the class-agnostic data set, we see that our method struggles to learn a good attention model. This is to be expected since there are many possible objects in both the train and test dataset. However, when we use the class-specific dataset with similar objects as in the EBAY dataset (cars, shoes, dresses, and pottery) results improve significantly. Our pixel-wise accuracy improves with 3\% over PLSA. On the image-wise evaluation we obtain even 86.59\% which is higher than any of the other methods which require a mask at test time. Our results of 87.27\% are obtained when we use the ground-truth segmentation mask as our attention map; note that all other methods (indicated with S) also use this mask at test time. As a comparison we also provide results with an image classification network; we use a pre-trained AlexNet and finetune on the training dataset. The results are more than 10\% lower than our method. 

Finally, we compare our testing results with human evaluation on EBAY. Humans are asked to choose the main color label for the object in each image; eight candidates without color blindness are asked to give one color label for each of 110 randomly chosen images from the EBAY dataset. We compute testing accuracy comparing to the ground truth of EBAY dataset and report the average accuracy ($88.98\%$) as the human evaluation baseline. This shows that our results have narrowed the gap with humans from 4\% to around 2\%. 

\subsection{Domain-Specific Color Naming}
The main objective of our paper is to provide a method which can be applied to new sets of domain-specific color names with only weakly labeled data. Here we evaluate our method on the four groups from our domain-specific dataset. We compare to the previously discussed classification network; the other methods cannot be applied in this setting.

Table~\ref{table:domain1} gives the results of color naming for the Eye, Lip and Horse color and Tomato growing stage. Our method outperforms the classification network on all groups. The attention network manages to identify the relevant objects as can be seen from the attention maps of some testing images in Fig. \ref{fig:domain_map}, where highlighted yellow regions indicate high-interest parts, and blue means low-interest parts. The smaller gains on the Horse and Tomato groups can be explained by the fact that the main object occupies most of the image and in that case the classification network also manages to extract the relevant color name. 

\section{Conclusions}
In this paper we have proposed a new network for the learning of domain-specific color names from weakly labeled data. This two-branch network learns, in an end-to-end fashion, a color name probability map for each pixel and an attention map. When joined, these maps result in a color name prediction for the image. Our method is the first color name method which does not require hand-labeled masks at testing time. Results show that the attention maps correctly identify the relevant image regions and that the network successfully learns domain-specific color names. In addition, we show that the pixel-wise and image-wise predictions of the network obtain state-of-the-art results on the EBAY dataset. 

\section*{Acknowledgement}
This work was supported by TIN2016-79717-R of the Spanish Ministry and the CERCA Programme / Generalitat de Catalunya, the EU Project CybSpeed MSCA-RISE-2017-777720, and Chinese National Natural Science Foundation under Grant 61603364. We also acknowledge the generous GPU support from Nvidia.

\bibliographystyle{IEEEtran}
\bibliography{refs}

\end{document}